\documentclass[journal]{IEEEtran}
\usepackage{amsmath,amsfonts}
\usepackage{algorithmic}
\usepackage{algorithm}
\usepackage{array}
\usepackage[caption=false,font=normalsize,labelfont=sf,textfont=sf]{subfig}
\usepackage{textcomp}
\usepackage{stfloats}
\usepackage{url}
\usepackage{verbatim}
\usepackage{graphicx}
\usepackage{cite}
\begin{document}

\title{Rethinking Soft Actor-Critic in High-Dimensional Action Spaces: The Cost of Ignoring Distribution Shift}

\author{Yanjun~Chen$^{1,2}$,~Xinming~Zhang$^{2}$,~Xianghui~Wang$^{2}$,~Zhiqiang~Xu$^{2}$,~Xiaoyu~Shen$^{2,*}$,~Wei~Zhang$^{2,*}$
\thanks{$^{*}$Xiaoyu Shen and Wei Zhang are joint corresponding authors (e-mail: xyshen@eitech.edu.cn, zhw@eitech.edu.cn).}
\thanks{$^{1}$Yanjun Chen is with the Department of Computing, The Hong Kong Polytechnic University, Hong Kong SAR, China (e-mail: yan-jun.chen@connect.polyu.hk).}%
\thanks{$^{2}$All authors are with the Digital Twin Institute, Eastern Institute of Technology, Ningbo, China.}
}






\maketitle

\begin{abstract}
Soft Actor-Critic algorithm is widely recognized for its robust performance across a range of deep reinforcement learning tasks, where it leverages the \textit{tanh} transformation to constrain actions within bounded limits. However, this transformation induces a distribution shift, distorting the original Gaussian action distribution and potentially leading the policy to select suboptimal actions, particularly in high-dimensional action spaces. In this paper, we conduct a comprehensive theoretical and empirical analysis of this distribution shift, deriving the precise probability density function (PDF) for actions following the \textit{tanh} transformation to clarify the misalignment introduced between the transformed distribution’s mode and the intended action output. We substantiate these theoretical insights through extensive experiments on high-dimensional tasks within the HumanoidBench benchmark. Our findings indicate that accounting for this distribution shift substantially enhances SAC’s performance, resulting in notable improvements in cumulative rewards, sample efficiency, and reliability across tasks. These results underscore a critical consideration for SAC and similar algorithms: addressing transformation-induced distribution shifts is essential to optimizing policy effectiveness in high-dimensional deep reinforcement learning environments, thereby expanding the robustness and applicability of SAC in complex control tasks.
\end{abstract}

\begin{IEEEkeywords}
Reinforcement learning, Deep learning, Machine learning, Probability distribution
\end{IEEEkeywords}

\section{Introduction}
Deep Reinforcement Learning (DRL) has become a cornerstone of modern artificial intelligence, driving transformative advancements across a wide range of domains, including autonomous systems, robotics, and strategic gaming. Leveraging reward-driven learning frameworks, DRL enables agents to develop optimal strategies for navigating complex and dynamic environments, establishing its efficacy particularly in continuous control tasks.

Among DRL algorithms, Soft Actor-Critic (SAC) algorithm~\cite{haarnoja2018soft} is widely recognized for its robust performance and unique blend of sample efficiency and stability. SAC integrates entropy maximization, which incentivizes exploration, with off-policy learning, which promotes efficient experience reuse. This design achieves rapid convergence and stable performance, positioning SAC as a state-of-the-art solution for continuous control tasks. SAC’s versatility has enabled its successful application in a wide range of fields, including robotic manipulation~\cite{kalashnikov2018scalable}, path-following control~\cite{ma2023sample}, robotic navigation~\cite{choi2021risk, wen2025deep}, robotic grasping~\cite{berscheid2021robot, moosmann2021separating, doi:10.1177/0278364917710318}, underwater glider sampling~\cite{zang2023energy}, and even medical domains such as optimizing ventilator settings~\cite{chen2022model}. These diverse applications underscore SAC’s flexibility and its capacity to deliver reliable performance across challenging and varied scenarios.

\begin{figure}[t]
\centering
\includegraphics[width=0.9\columnwidth]{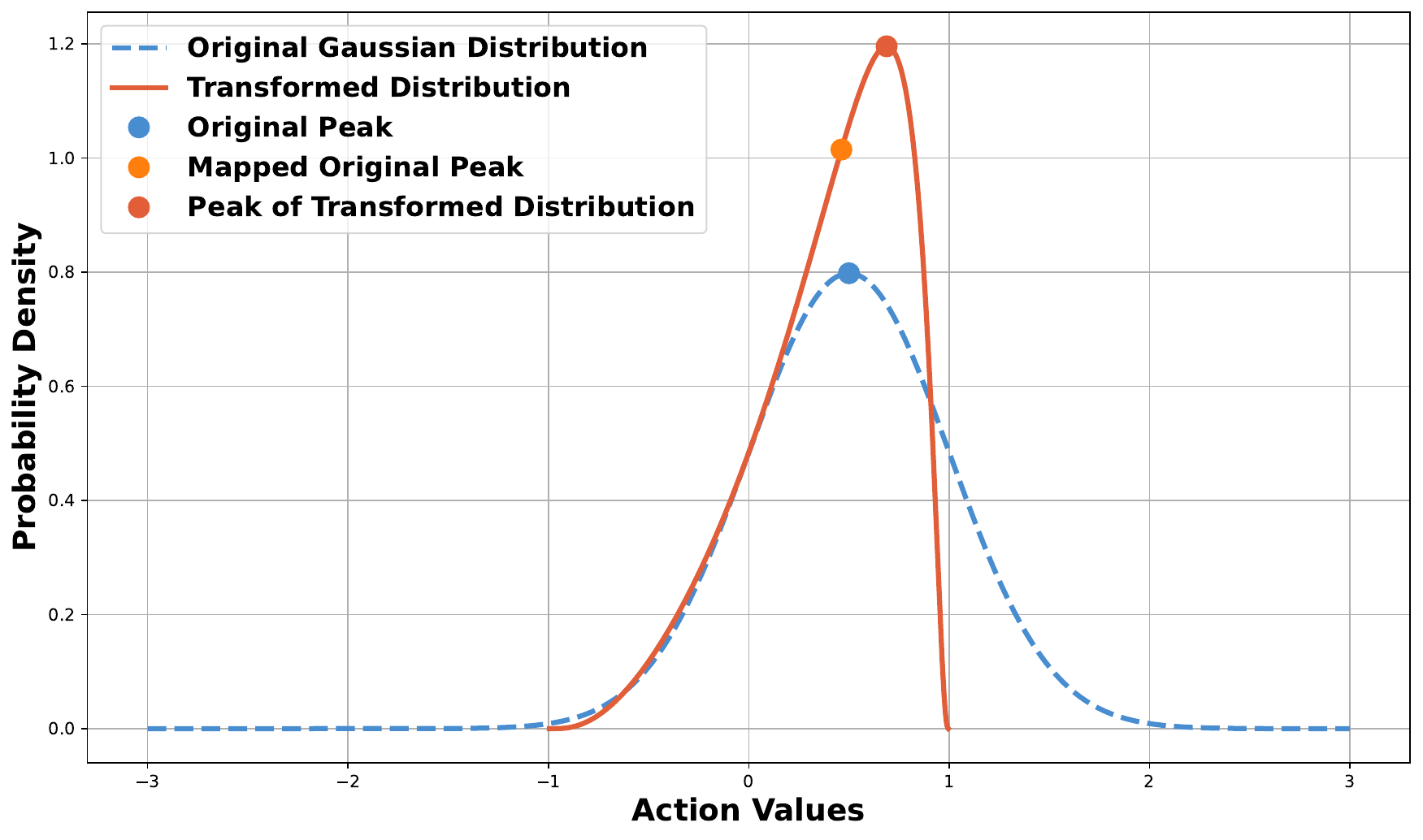}
\caption{Comparison of probability densities between the original Gaussian distribution and the transformed distribution after \textit{tanh} transformation.}
\label{fig:motivation}
\end{figure}

Despite the notable success of SAC, its action sampling mechanism is hindered by a significant yet underexplored limitation: the \textit{distribution shift} introduced by the \textit{tanh} transformation. This transformation, which is widely utilized to constrain actions within the bounded interval $[-1, 1]$, ensures compliance with environmental action constraints~\cite{haarnoja2018soft}. However, as a nonlinear operation, \textit{tanh} alters the original Gaussian distribution, inducing a systematic shift in probability densities. Specifically, the transformation compresses probability densities near the boundaries ($y \approx \pm 1$) while stretching them near the center ($y \approx 0$). This distortion creates a misalignment between the original and transformed action distributions, leading to biased action selection. The impact of this bias becomes particularly pronounced in high-dimensional action spaces, where the cumulative effect of distribution shifts across dimensions can significantly impair policy performance.

The issue of distribution distortion caused by nonlinear transformations is not unique to SAC or deep reinforcement learning. Similar challenges have been observed in variational autoencoders, where nonlinear transformations of latent variables lead to distribution shifts that degrade performance~\cite{kingma2013auto}. While action squashing functions like \textit{tanh} are commonly used in DRL algorithms~\cite{duan2016benchmarking}, prior research has predominantly focused on improving sample efficiency and algorithm stability~\cite{fujimoto2018addressing}, often neglecting the broader implications of distribution shifts. In SAC, the impact of this distortion becomes particularly pronounced in high-dimensional action spaces, where cumulative biases across dimensions exacerbate performance degradation in complex tasks, such as those involving humanoid robotics.

To bridge this gap, the present study conducts a comprehensive analysis of the distribution shift caused by the \textit{tanh} transformation in SAC. Specifically, we derive the precise probability density function (PDF) of the transformed actions, establishing a robust theoretical foundation for understanding how the \textit{tanh} transformation distorts the original Gaussian distribution. This mathematical framework clarifies the mechanisms by which distribution shifts impair policy optimization and action selection. As depicted in Fig.~\ref{fig:motivation}, the visualization emphasizes the divergence between the original Gaussian distribution and its \textit{tanh}-transformed counterpart, illustrating the practical implications of this shift on deep reinforcement learning performance.

To empirically validate the impact of distribution shift, we conducted experiments on the HumanoidBench benchmark~\cite{sferrazza2024humanoidbench}, a suite of high-dimensional, whole-body motion control tasks designed to emphasize the challenges of high-dimensional action spaces. These tasks offer an ideal platform for examining the cumulative effects of distribution shift. Leveraging standard cumulative reward metrics and advanced evaluation tools from the \texttt{rliable} library~\cite{agarwal2021deep}, we assessed the performance of standard SAC and a corrected SAC variant across key metrics, including Interquartile Mean (IQM), Median, and Sample Efficiency. These metrics provide a robust and comprehensive framework for quantifying the impact of distribution shift on policy performance across diverse conditions.

Our findings illuminate a critical limitation of SAC while offering broader implications for the design of deep reinforcement learning algorithms. By addressing the distribution shift caused by the \textit{tanh} transformation, we provide actionable insights to improve policy optimization in high-dimensional settings. These insights extend beyond SAC to other DRL algorithms employing similar transformations~\cite{schulman2017proximal, tang2017variational, schulman2015trust, wang2016sample}, with particular relevance to domains like humanoid robotics, where robust and reliable policies are essential for navigating complex, high-dimensional action spaces.

The main contributions of this paper are as follows:

\begin{itemize}
    \item \textbf{Identification of the distribution shift problem}: We identify and formalize the distribution shift caused by the \textit{tanh} transformation in SAC, highlighting its impact on action selection and policy optimization.
    
    \item \textbf{Mathematical derivation of the accurate probability density function}: We derive the precise PDF for \textit{tanh}-transformed actions, providing a theoretical framework to quantify the effects of distribution distortion.
    
    \item \textbf{Empirical validation of the impact on policy performance}: Using high-dimensional tasks from the HumanoidBench benchmark, we demonstrate that ignoring distribution shifts results in significant performance degradation, while correcting for these shifts improves robustness, reliability, and efficiency.
\end{itemize}

\section{Problem Identification}

\subsection{Action Sampling in SAC}

In SAC, the policy network generates actions by sampling from a Gaussian distribution to facilitate exploration. This action distribution is expressed as:

\begin{equation}
u \sim \mathcal{N}(\mu, \sigma^2)
\end{equation}
where $\mu$ and $\sigma^2$ denote the mean and variance predicted by the policy network. The Gaussian distribution, centered at $\mu$ with variability determined by $\sigma^2$, reflects the policy's uncertainty and enables exploration by promoting diversity in action selection.

\subsection{Action Constraints and Distribution Shift}

To ensure compliance with environment-specific action bounds, SAC employs the \textit{tanh} function to constrain actions within the interval $(-1, 1)$:

\begin{equation}
y = \tanh(u)
\end{equation}

While the \textit{tanh} transformation effectively bounds actions, its nonlinear nature distorts the original Gaussian distribution, introducing a \textit{distribution shift}. This shift results in an uneven probability density: it increases near the boundaries ($y \approx \pm 1$) and decreases near the center ($y \approx 0$), causing a non-uniform transformation of the action distribution.

\begin{figure}[t]
\centering
\includegraphics[width=0.9\columnwidth]{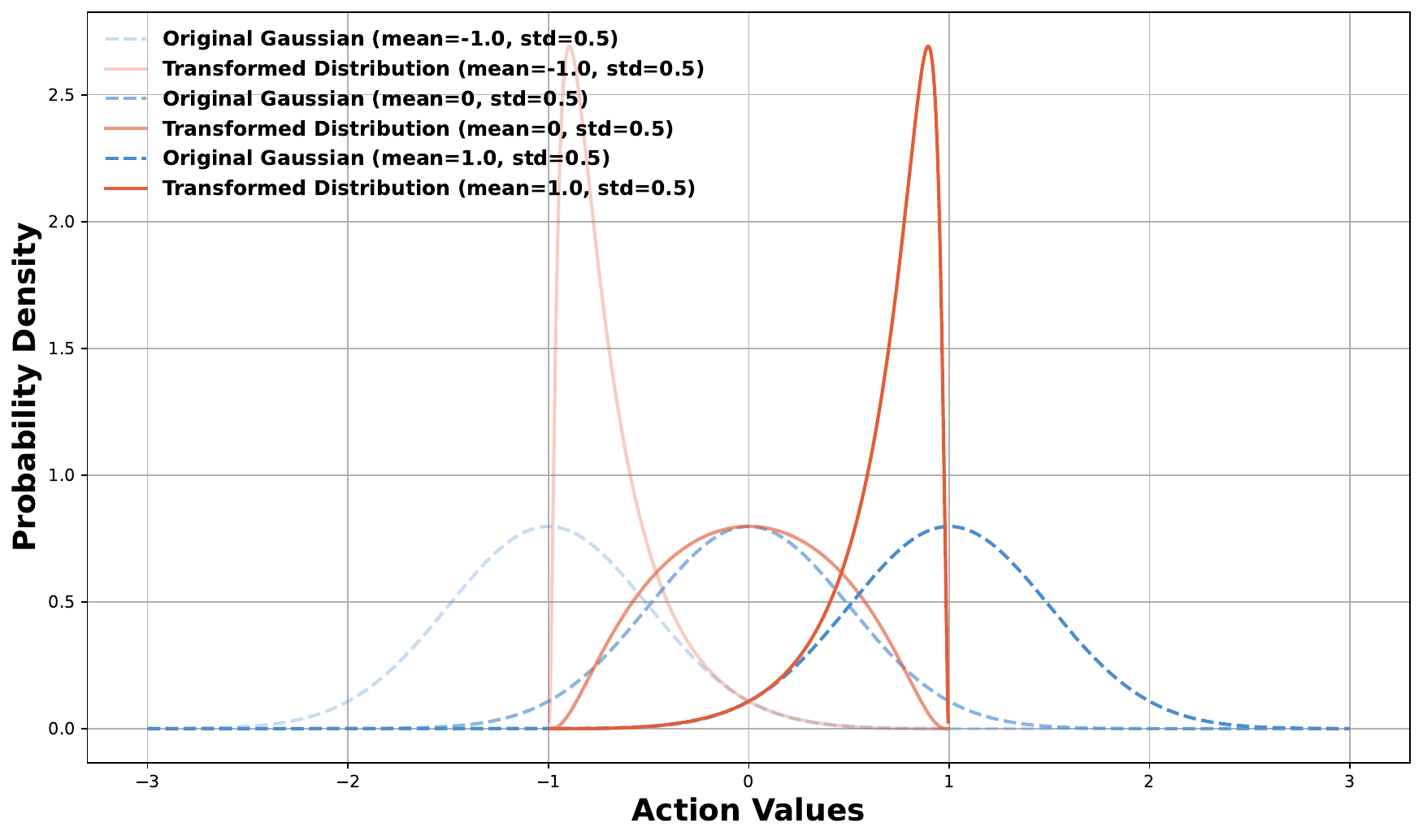}
\caption{Impact of the \textit{tanh} transformation on Gaussian distributions with different means and a fixed standard deviation (0.5). The figure highlights the variation in probability density introduced by the transformation.}
\label{fig:distribution_shift}
\end{figure}

The core mechanism of this distribution shift lies in the differential effect of \textit{tanh} across regions. Near the center ($y \approx 0$), the transformation “stretches” the distribution, reducing probability density, whereas near the boundaries ($y \approx \pm 1$), the transformation “compresses” the distribution, amplifying probability density. Fig.~\ref{fig:distribution_shift} illustrates this phenomenon by comparing Gaussian distributions with varying means (1.0, 0, -1.0) and a fixed standard deviation of 0.5. The blue dashed curves represent the original Gaussian distributions, while the red solid curves depict the distributions after applying the \textit{tanh} transformation.

\subsection{Action Bias from Distribution Shift}

In the standard SAC algorithm, actions are typically derived by directly applying the \textit{tanh} transformation to the mean output $\mu$ of the policy network:

\begin{equation}
y = \tanh(\mu)
\end{equation}

However, this approach overlooks the distribution shift induced by the \textit{tanh} transformation. Due to the nonlinear nature of \textit{tanh}, $\tanh(\mu)$ does not correspond to the mode (the highest-probability point) of the transformed distribution. Instead, the transformation introduces a \textit{bias}, leading to the selection of suboptimal actions during inference. In high-dimensional action spaces, such biases accumulate across dimensions, further exacerbating the degradation of policy performance.

To accurately represent the distribution after the transformation, it is necessary to apply the \textit{change of variables} formula, which incorporates the Jacobian determinant to correct the probability density. Starting with the original Gaussian distribution:

\begin{equation}
u \sim \mathcal{N}(\mu, \sigma^2)
\end{equation}

The probability density function (PDF) of the transformed distribution is obtained as:

\begin{equation}
p(y) = p(u) \left| \frac{du}{dy} \right|
\end{equation}

Here, $\left| \frac{du}{dy} \right|$ represents the Jacobian determinant, often referred to as the "stretching factor," introduced by the \textit{tanh} transformation. The resulting PDF of the transformed variable $y$ is expressed as:
\begin{equation}
\label{eq:pdf_y_full}
p(y) = \frac{1}{1 - y^2} \cdot \frac{1}{\sqrt{2\pi \sigma^2}} \exp\left( -\frac{\left( \mathrm{artanh}(y) - \mu \right)^2}{2\sigma^2} \right)
\end{equation}
where $\mathrm{artanh}(y) = \frac{1}{2} \ln \frac{1 + y}{1 - y}$. The term $\frac{1}{1 - y^2}$ captures the nonlinear distortion caused by the \textit{tanh} transformation. It amplifies probability density near the boundaries ($y \approx \pm 1$) and reduces it near the center ($y \approx 0$), thereby introducing a significant bias in the transformed distribution.

\subsection{Visualizing Action Bias}

\begin{figure}[t]
\centering
\includegraphics[width=0.9\columnwidth]{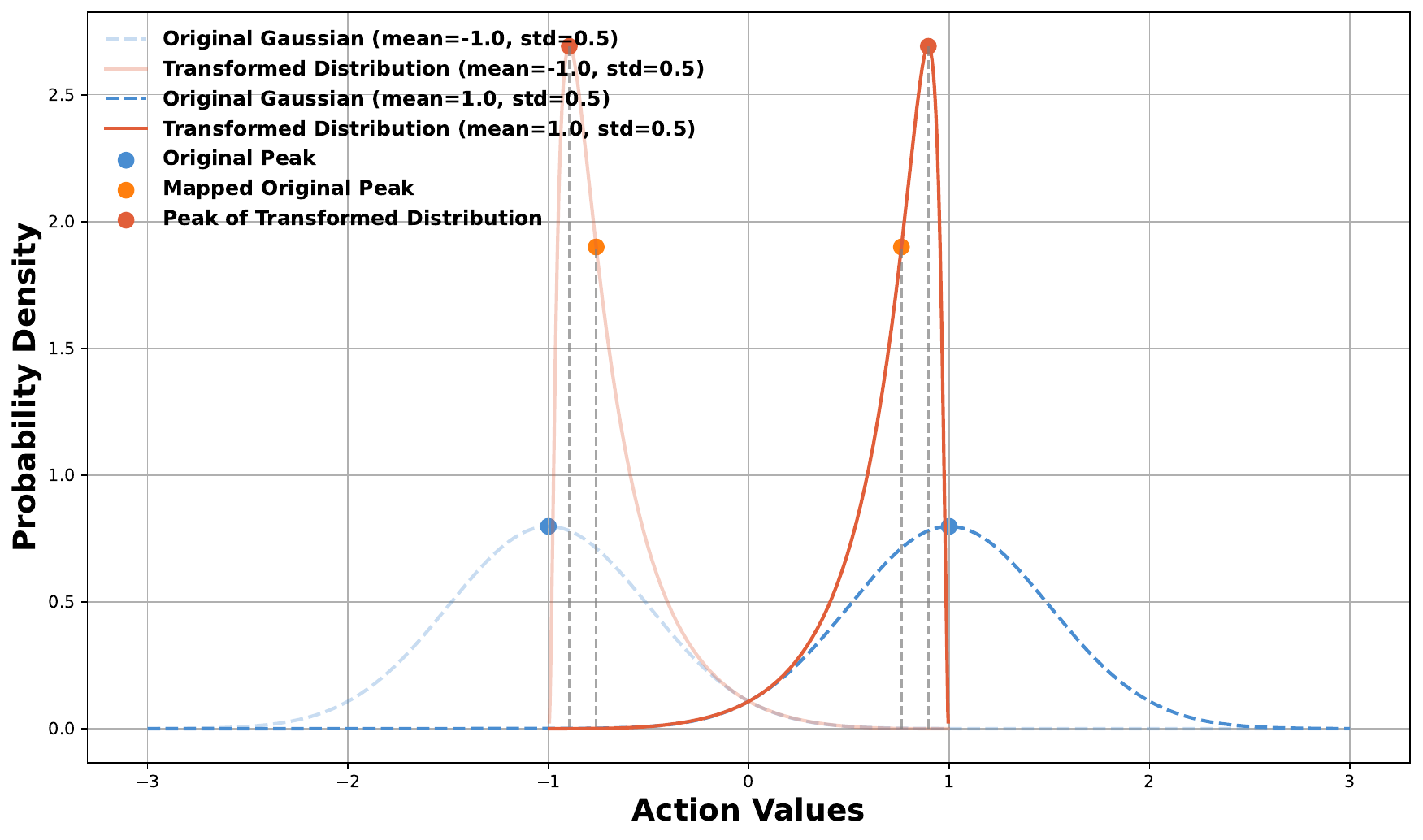}
\caption{Comparison of action selection with and without accounting for distribution shift. The orange point represents the transformed location of the original mode, while the red point indicates the mode of the \textit{tanh}-transformed distribution.}
\label{fig:action_bias}
\end{figure}

To visualize the action bias resulting from neglecting the distribution shift, Fig.~\ref{fig:action_bias} compares Gaussian distributions with means (1.0, -1.0) and a fixed standard deviation (0.5) before and after applying the \textit{tanh} transformation. The figure highlights three critical points:

\begin{itemize}
    \item The mode of the original Gaussian distribution (blue point),
    \item The mode of the transformed distribution (red point),
    \item The transformed location of the original mode (orange point).
\end{itemize}

This visualization underscores the misalignment introduced by the distribution shift, where the desired action (as indicated by the original mode) fails to align with the transformed output. This misalignment is particularly detrimental in high-dimensional action spaces, where cumulative distortions lead to substantial performance losses.

\section{Analysis Methodology}


\subsection{Theoretical Analysis of High-Dimensional Spaces}

The distortion introduced by the \textit{tanh} transformation in action sampling is significantly exacerbated in high-dimensional action spaces. In these settings, each dimension independently contributes to the overall distribution shift, resulting in a cumulative effect that amplifies misalignments between the intended and actual action distributions.

For a single action dimension \( u \sim \mathcal{N}(\mu, \sigma^2) \), the \textit{tanh} transformation skews the original Gaussian distribution, as mathematically characterized in Equation~\ref{eq:pdf_y_full}. In a $d$-dimensional action space, the cumulative distortion scales approximately with the dimensionality \( d \), given that each dimension is subject to the same nonlinear transformation. Tasks involving high-dimensional control, such as humanoid robotics, are particularly susceptible to this phenomenon, as they require coordinated precision across numerous degrees of freedom.

\begin{figure}[t]
\centering
\includegraphics[width=0.9\columnwidth]{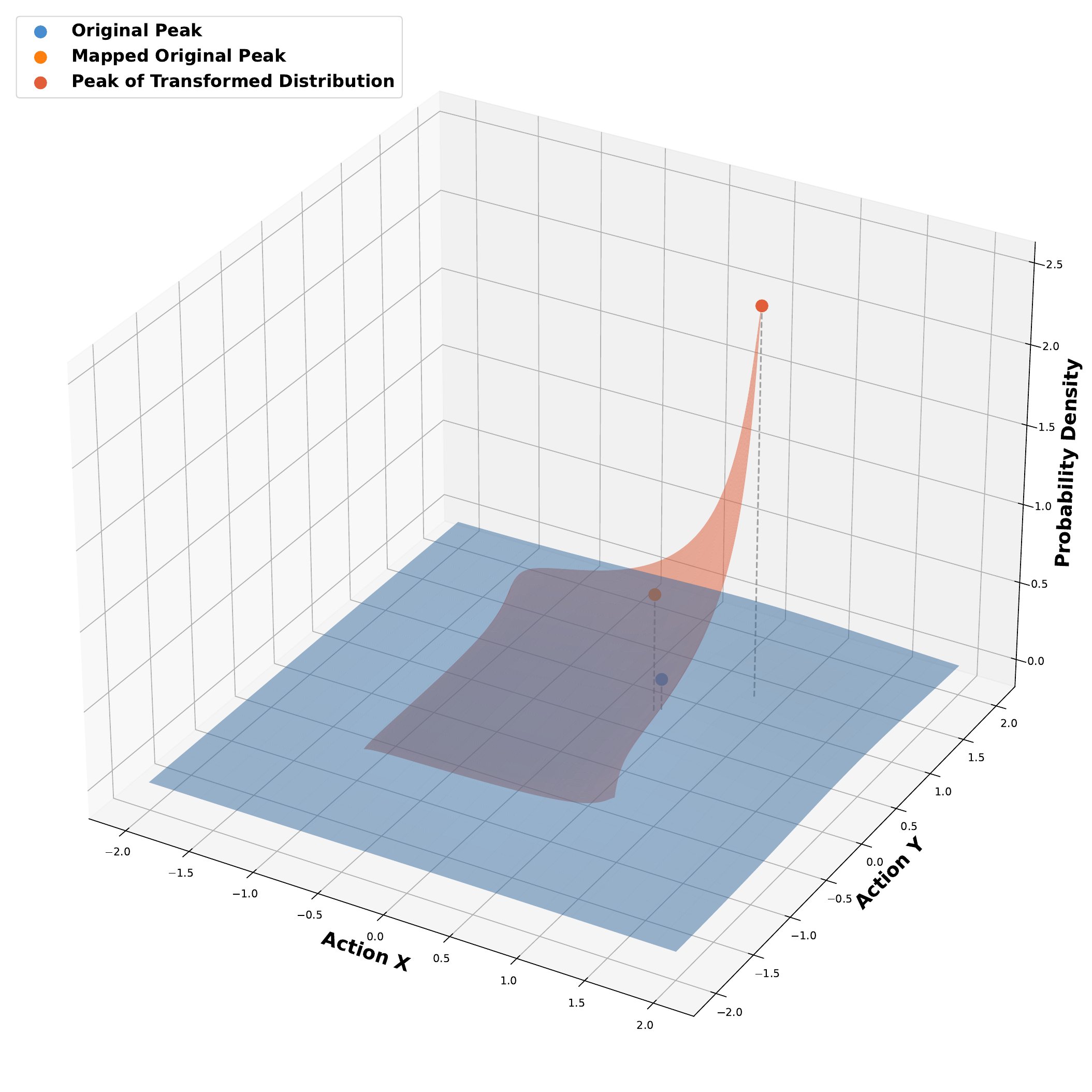}
\caption{Illustration of the distribution shift in a 2D action space induced by the \textit{tanh} transformation. Each dimension contributes independently to the overall misalignment, amplifying the effect in high-dimensional action spaces.}
\label{fig:action_selection_bias}
\end{figure}

To illustrate this effect, Fig.~\ref{fig:action_selection_bias} provides an example of the distribution shift in a two-dimensional action space. Although limited to two dimensions for visualization purposes, this figure effectively demonstrates the core mechanics of the compounded distortion that occurs in higher-dimensional spaces. The visualization highlights the divergence between the transformed mean and the mode of the resulting distribution, which increases with the number of action dimensions. This cumulative misalignment directly impacts policy performance, particularly in tasks requiring fine-grained control.

\subsection{Experimental Design and Setup}

To validate the theoretical insights, we designed a series of experiments to evaluate the effects of distribution shift on SAC performance. These experiments leveraged tasks from the HumanoidBench benchmark, a platform specifically suited for high-dimensional action space evaluations. This section details the experimental design, including the environment, algorithmic variants, evaluation metrics, and hyperparameter configurations.

\subsubsection{Environment}

HumanoidBench~\cite{sferrazza2024humanoidbench} provides a comprehensive suite of tasks tailored for testing deep reinforcement learning algorithms in high-dimensional action spaces, making it an ideal benchmark for this study. Each task operates within a normalized action space \([-1, 1]^{\|A\|}\), where \( \|A\| = 61 \) denotes the total number of action dimensions: 19 for the humanoid body and 21 for each hand. This high-dimensional setup allows for systematic exploration of how distribution shifts accumulate across dimensions and impact policy performance.

The benchmark comprises four diverse tasks, each emphasizing distinct aspects of whole-body coordination and dexterous manipulation:

\begin{itemize}
    \item \textbf{\textit{powerlift}}: Evaluates whole-body coordination and force control by requiring the humanoid to lift a barbell-shaped object of specified mass.
    \item \textbf{\textit{cube}}: Focuses on dexterous hand manipulation, challenging the humanoid to rotate two in-hand cubes to achieve randomly assigned target orientations.
    \item \textbf{\textit{bookshelf\_simple}}: Tests precision and planning by requiring sequential picking and placing of items across specified shelves, following a fixed order.
    \item \textbf{\textit{bookshelf\_hard}}: Increases the complexity of \textit{bookshelf\_simple} by randomizing the order of items and destinations, demanding adaptability and robustness.
\end{itemize}

\begin{figure*}[t]
\centering
\includegraphics[width=\textwidth]{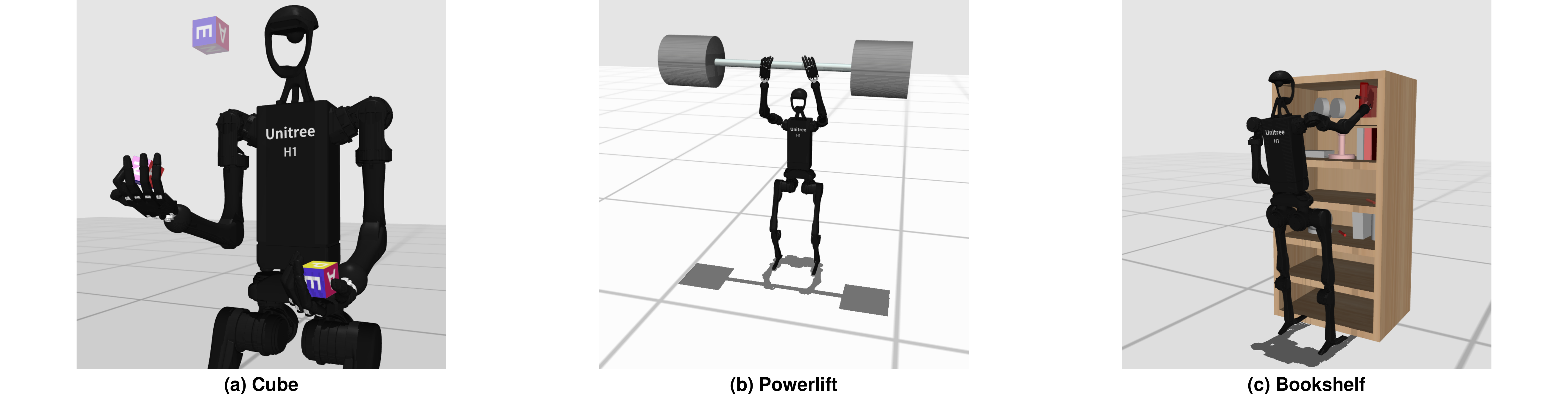}
\caption{Overview of the tasks from the HumanoidBench benchmark: (a) \textit{cube}, (b) \textit{powerlift}, and (c) \textit{bookshelf}.}
\label{fig:tasks_overview}
\end{figure*}

Fig.~\ref{fig:tasks_overview} provides a visual representation of these tasks, showcasing their diversity and complexity.

\subsubsection{Algorithms}

To assess the impact of distribution shift, we compared two algorithmic variants of SAC:

\begin{itemize}
    \item \textbf{Standard SAC}: This baseline variant directly applies the \textit{tanh} transformation to the mean action output, without considering the resulting distribution shift. It represents the conventional approach widely used in deep reinforcement learning.
    \item \textbf{Corrected SAC}: This variant incorporates adjustments to account for distribution shift during inference by aligning actions with the mode of the transformed distribution. The adjustment is implemented through a sampling process to identify the most probable action \( y^* \), as detailed in Algorithm~\ref{alg:optimal_action_sampling_inference}.
\end{itemize}

\begin{algorithm}[t]
\small
\caption{Corrected SAC Sampling}
\label{alg:optimal_action_sampling_inference}
\textbf{Input}: Mean \( \mu \) and variance \( \sigma^2 \) from the policy network.\\
\textbf{Output}: Most probable action \( y^* \).
\begin{algorithmic}[1]
    \STATE Initialize \( y_0 = -0.999 \), \( \Delta y = 0.0004 \), \( N = 5000 \).
    \STATE Define \( p(y) \) using Eq.~\ref{eq:pdf_y_full}.
    \FOR{ \( i = 0 \) to \( N \)}
        \STATE \( y_i = y_0 + \Delta y \cdot i \).
        \STATE Calculate \( p(y_i) \).
    \ENDFOR
    \STATE \( y^* = \arg \max_{y_i} p(y_i) \).
    \STATE \textbf{return} \( y^* \).
\end{algorithmic}
\end{algorithm}

\subsubsection{Evaluation Metrics}

To ensure comprehensive and statistically reliable comparisons, we utilized a combination of general and \texttt{rliable} metrics~\cite{agarwal2021deep}, tailored for robust deep reinforcement learning evaluation. The selected metrics are as follows:

\begin{itemize}
    \item \textbf{Cumulative Rewards}: A widely used metric that measures the total rewards accumulated by an agent over an episode. This general metric serves as a primary indicator of overall policy performance, providing a straightforward yet powerful means to compare algorithm efficacy.
    \item \textbf{Interquartile Mean (IQM)}: A robust statistic computed from the middle 50\% of scores, reducing sensitivity to outliers and delivering a reliable representation of central performance tendencies.
    \item \textbf{Median}: Represents the 50th percentile of scores, offering a robust measure of central tendency that remains unaffected by extreme values or skewed distributions.
    \item \textbf{Mean}: The arithmetic average of all scores, capturing overall performance trends but susceptible to the influence of outliers.
    \item \textbf{Performance Profiles}: Illustrates the proportion of runs achieving or surpassing specified performance thresholds, providing a comprehensive view of algorithm reliability and robustness, especially under challenging conditions.
    \item \textbf{Sample Efficiency}: Evaluates the rate at which each algorithm achieves stable performance over training frames, offering insights into learning speed and convergence stability.
\end{itemize}

While the \texttt{rliable} metrics—including IQM, Median, Performance Profiles, and Sample Efficiency—offer statistically robust tools for reliable comparisons across runs, Cumulative Rewards serves as an intuitive and widely recognized benchmark for evaluating deep reinforcement learning performance. Combined with stratified bootstrap confidence intervals, these metrics provide a comprehensive and balanced framework for assessing the performance of Standard SAC and Corrected SAC across diverse task conditions.

\subsubsection{Hyperparameter Settings}

To ensure consistency and reproducibility, the experiments strictly followed the hyperparameter configurations specified in \cite{sferrazza2024humanoidbench}. General training parameters, including seed range, batch size, and evaluation intervals, are detailed in Table~\ref{tab:general_hyperparameters}, while SAC-specific settings, such as actor and critic learning rates, target entropy, and $\alpha$, are outlined in Table~\ref{tab:sac_hyperparameters}.

\begin{table}[t]
\renewcommand{\arraystretch}{1.3}
\setlength{\tabcolsep}{22pt}
\caption{General hyperparameter settings for the experiments, following \cite{sferrazza2024humanoidbench}.}
\centering
\label{tab:general_hyperparameters}
\begin{tabular}{lc}
\hline
\hline
\textbf{General Hyperparameters} & \textbf{Value} \\
\hline
Seed range                 & 0 -- 4           \\
Evaluation episodes        & 10               \\
Evaluation interval        & 10,000           \\
Batch size                 & 64               \\
Maximum steps              & 10,000,000       \\
Start steps                & 10,000           \\
\hline
\hline
\end{tabular}
\end{table}

\begin{table}[t]
\renewcommand{\arraystretch}{1.3}
\setlength{\tabcolsep}{22pt}
\caption{SAC-specific hyperparameter configurations for the experiments, following \cite{sferrazza2024humanoidbench}.}
\centering
\label{tab:sac_hyperparameters}
\begin{tabular}{lc}
\hline
\hline
\textbf{SAC-Specific Configurations} & \textbf{Value} \\
\hline
Actor learning rate        & 0.0003           \\
Critic learning rate       & 0.0003           \\
Temperature learning rate  & 0.0003           \\
Hidden dimensions          & [256, 256]       \\
Discount factor            & 0.99             \\
$\tau$                     & 0.005            \\
Target update rate         & 0.005            \\
$\alpha$                   & Auto             \\
\hline
\hline
\end{tabular}
\end{table}

\section{Experimental Results}


\subsection{Results}

The experimental results on HumanoidBench tasks highlight the advantages of distribution-aware action selection in SAC across key metrics: cumulative rewards, reliability, robustness, and sample efficiency.

\subsubsection{Cumulative Rewards}

\begin{figure}[t]
\centering
\includegraphics[width=1\columnwidth]{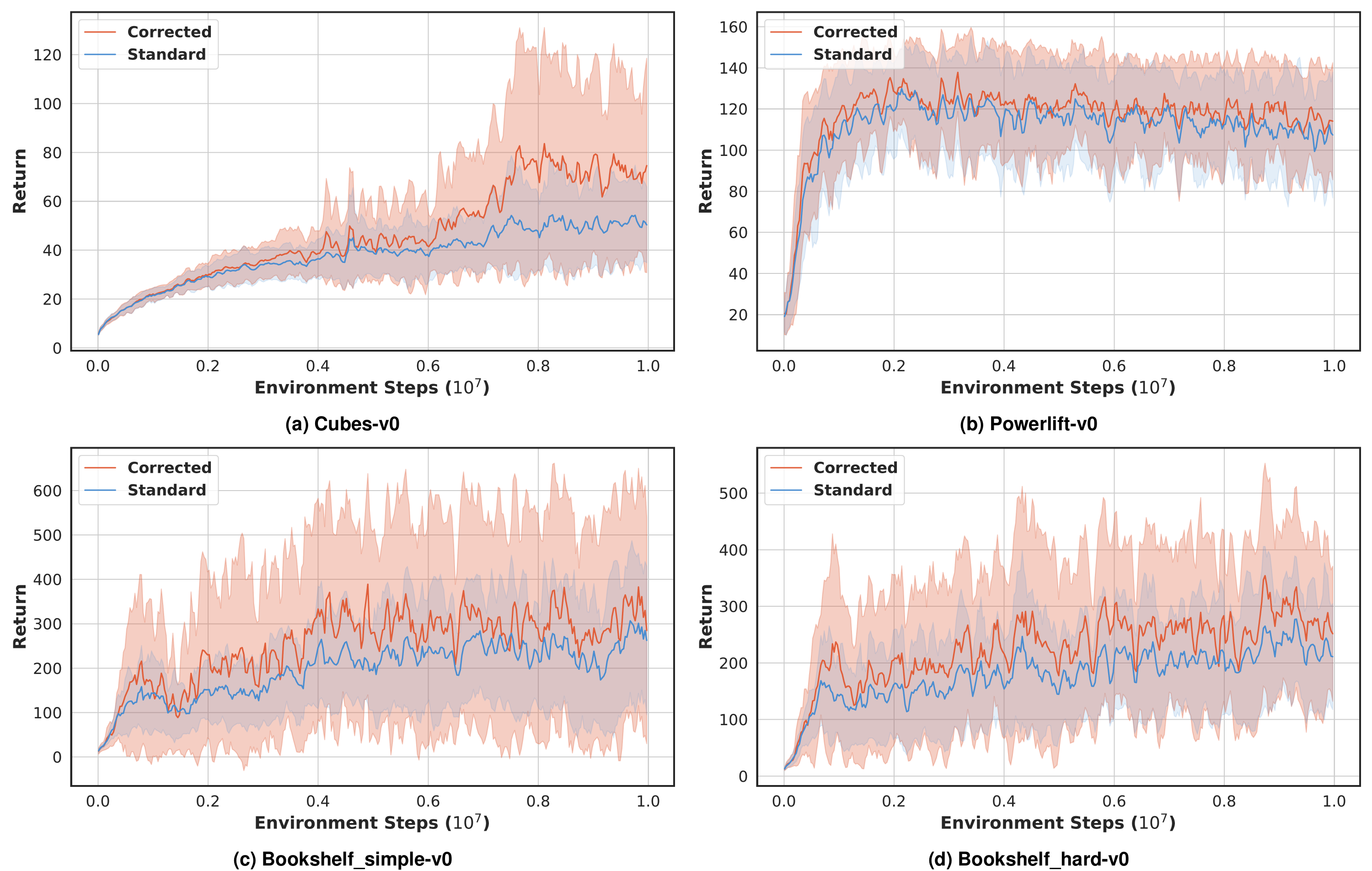}
\caption{Comparison of average cumulative rewards between Standard SAC and Corrected SAC across four tasks in HumanoidBench. Corrected SAC consistently achieves higher rewards, demonstrating the benefits of addressing distribution shift.}
\label{fig:cumulative_rewards}
\end{figure}

Corrected SAC consistently outperforms Standard SAC in cumulative rewards across all tasks, particularly in high-dimensional environments where compounded distribution shifts degrade Standard SAC’s performance. Fig.~\ref{fig:cumulative_rewards} illustrates the significant gains achieved by Corrected SAC.

\subsubsection{Reliability and Robustness}

\begin{figure}[t]
\centering
\includegraphics[width=1\columnwidth]{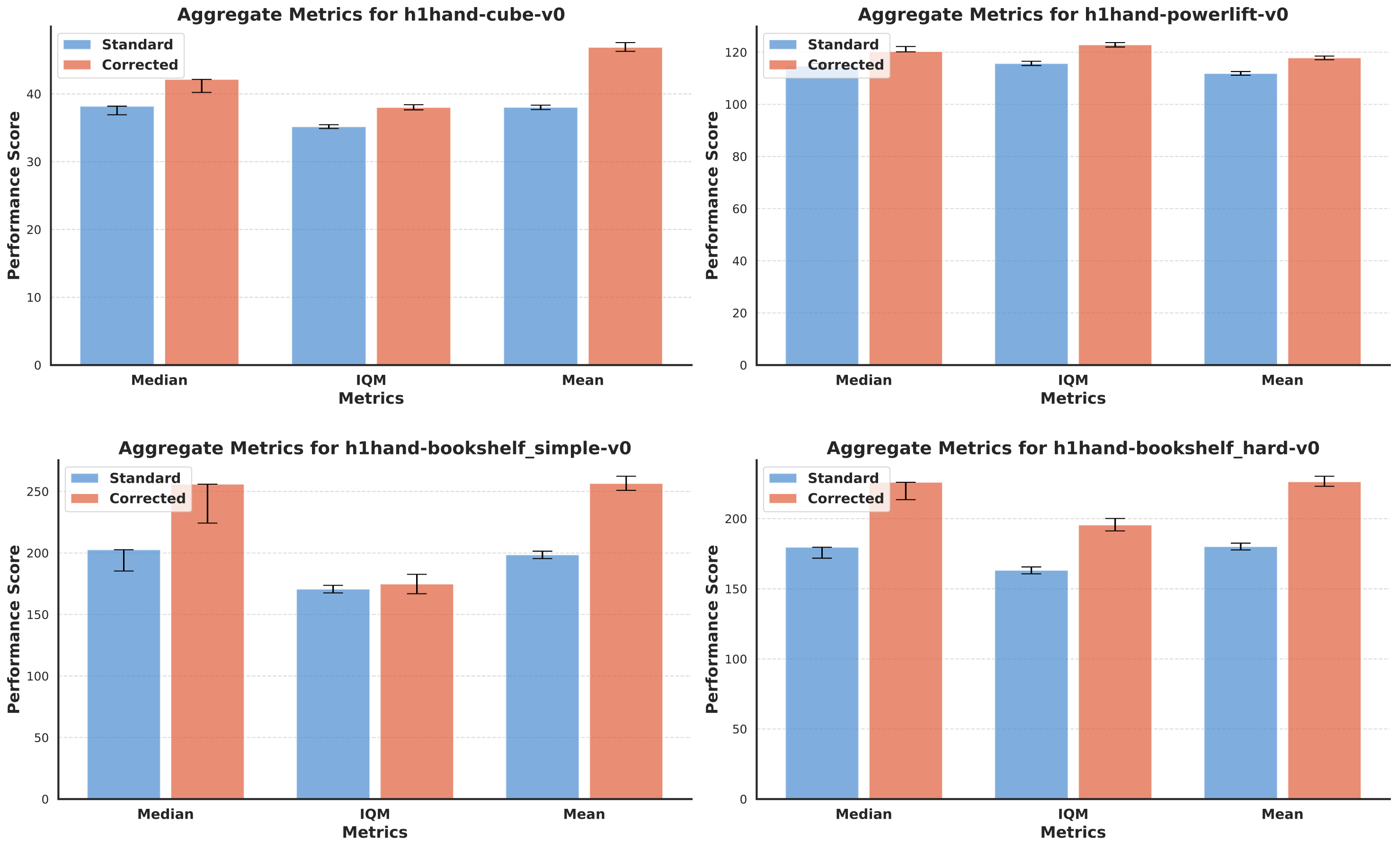}
\caption{Aggregate metric comparisons (IQM, Median, and Mean) between Standard SAC and Corrected SAC across tasks. Corrected SAC consistently outperforms Standard SAC across all robust performance metrics.}
\label{fig:aggregate_metrics}
\end{figure}

Metrics including Interquartile Mean (IQM), Median, and Mean highlight Corrected SAC’s enhanced reliability and robustness. Fig.~\ref{fig:aggregate_metrics} demonstrates its superior performance, reflecting better stability and generalization across diverse tasks.

\subsubsection{Performance Profiles}

\begin{figure}[t]
\centering
\includegraphics[width=1\columnwidth]{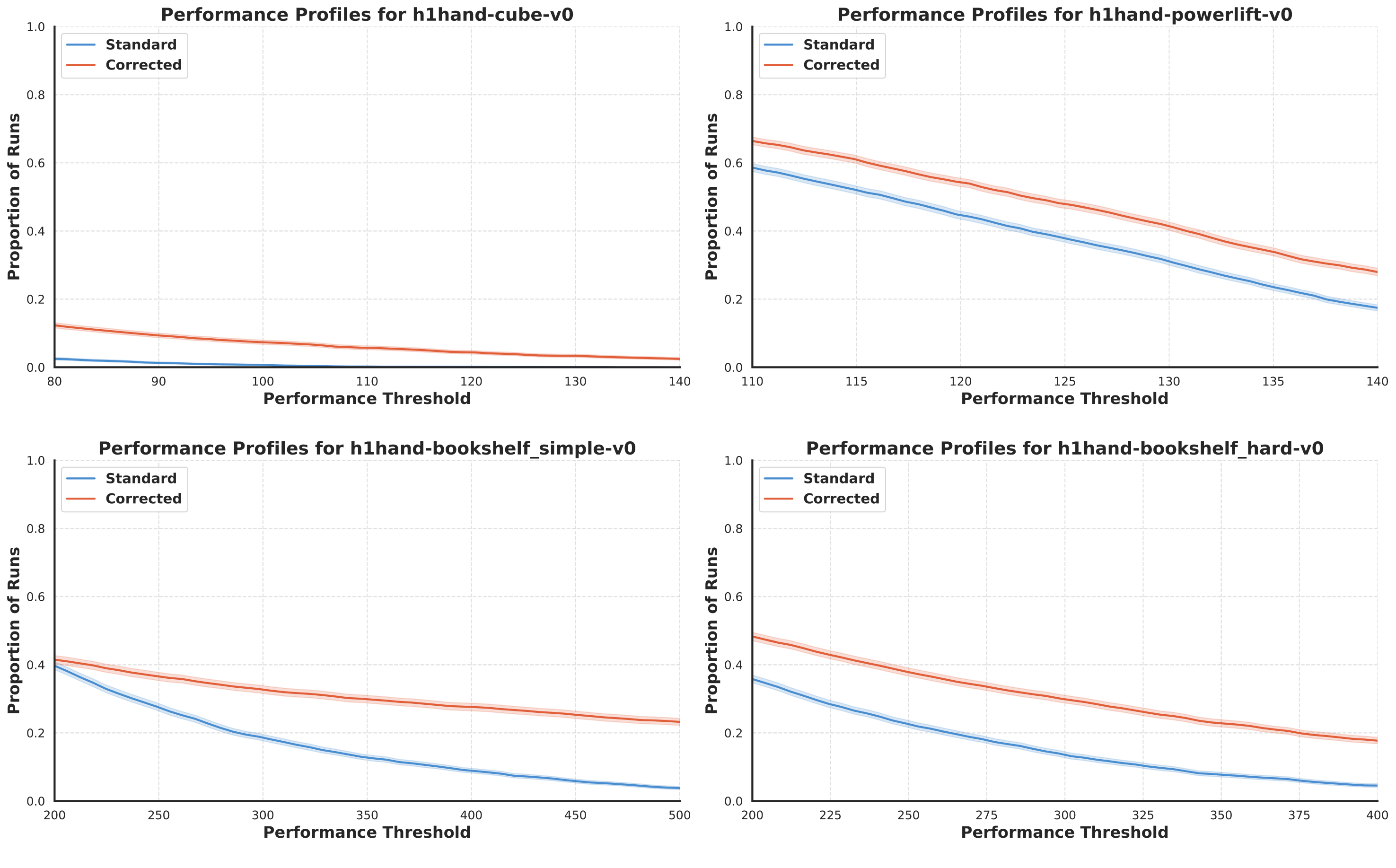}
\caption{Performance profiles of Standard SAC and Corrected SAC across tasks, illustrating the proportion of runs achieving specified performance thresholds. Corrected SAC consistently outperforms Standard SAC at higher performance levels.}
\label{fig:performance_profiles}
\end{figure}

Performance profiles, which capture the proportion of runs exceeding performance thresholds, provide a comprehensive view of algorithm reliability. As shown in Fig.~\ref{fig:performance_profiles}, Corrected SAC achieves consistently higher proportions, especially in more demanding scenarios.

\subsubsection{Sample Efficiency}

\begin{figure}[t]
\centering
\includegraphics[width=1\columnwidth]{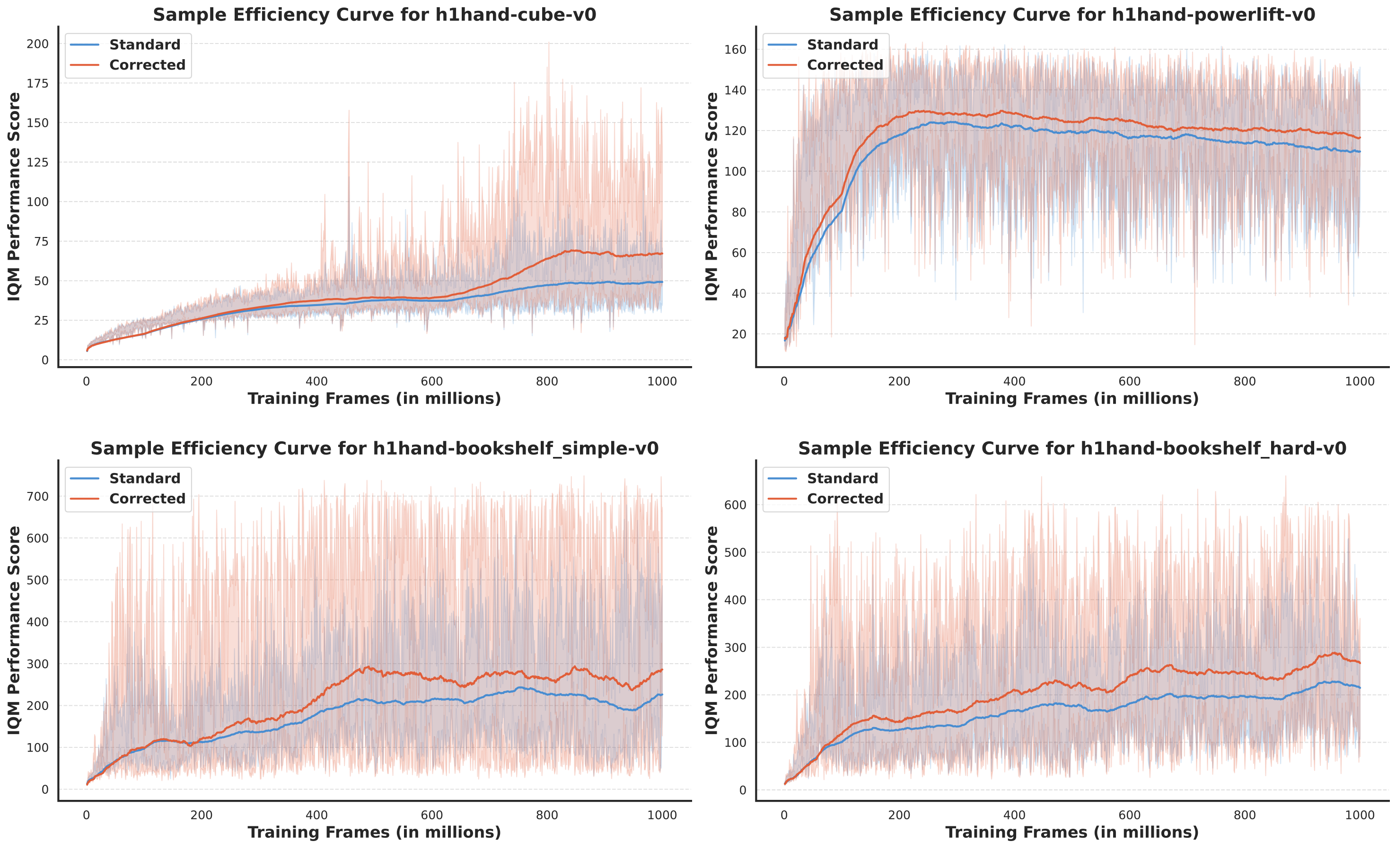}
\caption{Sample efficiency curves comparing Standard SAC and Corrected SAC across tasks. Corrected SAC achieves faster and more reliable performance improvements, demonstrating enhanced learning efficiency.}
\label{fig:sample_efficiency}
\end{figure}

Corrected SAC demonstrates superior sample efficiency, achieving stable performance with fewer training frames. Fig.~\ref{fig:sample_efficiency} highlights its faster convergence and more stable learning, underscoring the value of distribution-aware adjustments in high-dimensional settings.

\subsection{Discussion}

The experimental results reveal the significant impact of distribution shift on SAC’s performance, particularly in high-dimensional action spaces. While the \textit{tanh} transformation is essential for bounding actions within permissible ranges, its nonlinear nature introduces distortions in the action distribution. This distortion leads to a misalignment between the transformed action outputs and the policy’s intended actions, causing suboptimal action selection during inference.

Corrected SAC addresses this issue by aligning action selection with the mode of the transformed distribution, effectively mitigating the effects of distribution shift. This approach results in substantial improvements across key metrics, including cumulative rewards, reliability, robustness, and sample efficiency, as evidenced by our findings. These results underscore the critical importance of distribution-aware action selection, particularly for high-dimensional and complex environments such as humanoid robotics.

This study has broader implications beyond SAC, highlighting the relevance of addressing transformation-induced biases in deep reinforcement learning algorithms that employ nonlinear transformations like \textit{tanh} to constrain action spaces. Correcting such biases could yield similar benefits across other algorithms and applications. Future research should explore the generalization of these insights to additional continuous control methods, enabling more reliable and robust deep reinforcement learning solutions across diverse tasks and environments.

\section{Conclusion}

This study comprehensively analyzed the distribution shift introduced by the \textit{tanh} transformation in SAC algorithm, focusing on its detrimental impact in high-dimensional action spaces. While the \textit{tanh} transformation effectively constrains actions within bounded ranges, its nonlinear nature distorts the original Gaussian distribution, resulting in a misalignment that degrades action selection and policy performance. Through rigorous theoretical derivations and extensive empirical validation on the HumanoidBench benchmark, we quantified this degradation and demonstrated that aligning action selection with the mode of the transformed distribution significantly enhances SAC’s performance. The proposed correction method yielded substantial improvements in cumulative rewards, sample efficiency, and robustness, particularly in high-dimensional scenarios where distribution shifts are amplified. These findings underscore the importance of addressing transformation-induced biases in SAC and similar deep reinforcement learning algorithms to ensure optimal policy reliability and effectiveness in complex environments. Beyond SAC, this work highlights broader implications for deep reinforcement learning algorithm design, emphasizing the necessity of accounting for distribution shifts to enhance performance across diverse applications, such as humanoid robotics and other high-dimensional continuous control tasks.


\bibliographystyle{IEEEtran}
\bibliography{IEEEabrv}

\end{document}